# `FWDA`: a Fast Wishart Discriminant Analysis with its Application to Electronic Health Records Data Classification

Haoyi Xiong, Wei Cheng, Wenqing Hu, Jiang Bian, and Zhishan Guo

April 25, 2017

## Abstract

Linear Discriminant Analysis (LDA) on Electronic Health Records (EHR) data is widely-used for early detection of diseases. Classical LDA for EHR data classification, however, suffers from two handicaps: the *ill-posed* estimation of LDA parameters (e.g., covariance matrix), and the "linear inseparability" of EHR data. To handle these two issues, in this paper, we propose a novel classifier `FWDA` — Fast Wishart Discriminant Analysis, that makes predictions in an ensemble way. Specifically, `FWDA` first *surrogates* the distribution of inverse covariance matrices using a Wishart distribution estimated from the training data, then "weighted-averages" the classification results of multiple LDA classifiers parameterized by the sampled inverse covariance matrices via a Bayesian Voting scheme. The weights for voting are optimally updated to adapt each new input data, so as to enable the nonlinear classification. Theoretical analysis indicates that `FWDA` possesses a fast convergence rate and a robust performance on high dimensional data. Extensive experiments on large-scale EHR dataset show that our approach outperforms state-of-the-art algorithms by a large margin.

## 1 Introduction

The ubiquity of Electronic Health Records (EHR) [1, 2] in healthcare systems provides an unique opportunity for early detection of patients' potential diseases using their historical health records. Existing researches on it first extract useful features, such as diagnosis-frequencies [1, 3, 2], pairwise diagnosis transition [4, 5], and graphs of diagnosis sequences [6], to represent each patient's EHR data using the representation learning techniques. Then, supervised learning techniques are adopted to train predictive models, such as Support Vector Machine (SVM), Random Forest (RF), Bayesian Network, and Linear Discriminant Analysis (LDA) [1, 3, 2, 4, 7].

Among these methods, LDA is frequently used as one of the common performance benchmarks [4, 7], because of LDA's provable bayesian optimality [8]. However, recent studies demonstrate the limitation of LDA under high dimension low sample size (HDLSS) settings [9], such as the EHR records [10]. Because it is difficult to recover the "true" parameters, e.g., covariance matrix, from a *relatively small* number of training samples. When the number of dimensions of EHR data is larger than the number of samples, the sample covariance estimation used in classical LDA, is singular and not invertible. In this case, LDA cannot produce any valid prediction. Even when the sample size is larger than the number of dimensions, the sample (inverse) covariance estimation could be quite different with the "true" (inverse) covariance matrix, with an inconsistent estimate of the largest eigenvalues and almost-orthogonal eigenvectors to the truth [11]. Such *ill-posed estimation problem* significantly degrades the performance of LDA. Moreover, EHR data is usually not linearly separable [1, 2].

To address the ill-posed problems and the linear inseparability of the data, several regularization-based methods have been proposed to accurately estimate the (inverse) covariance matrix [12, 13, 14] or linear coefficients [15, 16] under high dimension and low sample size settings [17]. Further, to handle the non-linearity, some kernel-based or nonparameteric LDA classifiers [18, 19, 20, 21] have been proposed. In summary, these methods intend to improve LDA classification through optimizing the parameters of LDA, such as (inverse) covariance matrices, linear projection metrics, or kernel settings, in a so-called *optimal model selection* manner [22].

Instead of "bidding" the optimal parameter in the full and usually unknown parameter space, in this work, we intend to improve LDA in an ensemble way [23], while *adapting to the new input data*. Specifically, we first sample a set of (inverse) covariance matrices from the *both training data and the new input data*, then "weighted-averages" the classification results of multiple LDA classifiers parameterized by the sampled inverse covariance matrices via a *Bayesian Voting Scheme* [24]. Theoretical studies show that such Bayesian voting scheme can secure a *wider margin* and guarantee a good classification performance with a *lower generalization error bound* [24]. This theoretically guarantees that the proposed framework can "on average" outperform those regularization-based LDA classifiers using only *single* (inverse) covariance matrix estimator [25]. More importantly, the sampled (inverse) covariance matrices used by different LDA classifiers are updated with each new input data instance. In this way, the proposed classifier enables nonlinear classification by leveraging local information of the input data.

However, the aforementioned *Input-Adaptive Bayesian Voting Scheme* is not computationally efficient. As the sampled (inverse) covariance matrices are assumed to be updated to adapt each new input data for classification, the sampling complexity is very high. Especially, when the number of dimensions of data is high, it is quite time-consuming to sample the (inverse) covariance matrices, while ensuring each sampled matrix is positive-semidefinite. Thus, we propose a novel method FWDA – Fast Wishart Discriminant Analysis, which can approximate the optimal prediction results with minimal sampling efforts.

Specifically, FWDA first *surrogates* the distribution of inverse covariance matrices using a Wishart distribution estimated from the training data, then a set of inverse covariance matrices are sampled based on the distribution. The "weighted-averaged" result over the classification results from LDA classifiers parameterized by these sampled inverse covariance matrices are used for prediction. The "weights" are updated by each new input data for classification optimally in a Bayes manner. In this way, FWDA can approximate to the aforementioned *input-adaptive Bayesian voting* schema, with proven convergence rate. Our theoretical analysis further proves that (1) the error of approximation could quickly converge with the increasing number of sampled inverse covariance matrices $m$ in speed $\mathcal{O}(m^{-\frac{1}{2}})$; and (2) the error is not sensitive to the dimensions of the data, that means the performance of high dimensional data classification could be well-guaranteed.

In the rest of the paper, we first introduce the backgrounds, then we formulate the problem of research and elaborate the technical challenges in Section 2. In Section 3, we present the proposed algorithm FWDA, with the theoretical analysis on the approximation performance. In Section 4, we evaluate FWDA with other baseline algorithms for early detection of diseases using large-scale real EHR data. The results show that FWDA significantly outperforms baseline algorithms by a large margin.

## 2 Background and Problem Formulation

In this section, we first introduce the preliminaries of our research, then formulate the research problem of this paper.

### 2.1 Binary Classification for Early Detection of Diseases using EHR data

First of all, we introduce the EHR data representation using diagnosis-frequency vectors, and present settings of disease detection through binary classification of diagnosis-frequency vectors. Later, we briefly discuss the solution based on the typical LDA classifier.

*EHR Data Representation using Diagnosis-Frequency Vectors -* There are many existing approaches to represent EHR data including the use of diagnosis-frequencies [1, 3, 2], pairwise diagnosis transition [4, 5], and graph representations of diagnosis sequences [6]. Among these approaches, the diagnosis-frequency is a common way to represent EHR data.

Given each patient's EHR data, this method first retrieves the diagnosis codes [26] recorded during each visit. Next, the frequency of each diagnosis appearing in all past visits are counted, followed by further transformation on the frequency of each diagnosis into a vector of frequencies. For example, $\langle 1, 0, \ldots, 3 \rangle$, where 0 means the second diagnosis does not exist in all past visits. In this paper, we denote the dimension of diagnosis-frequency vectors as $p$. Note that the dimension $p$ of original codes is usually larger than $15,000$. Even using clustered codes, $p$ is usually larger than 250 [27], while the number of samples for training $n$ is frequently smaller than $p$.

*Early Detection by Binary Classification -* Given $m$ training samples (i.e., EHR frequency vectors) along with corresponding labels i.e., $(x_1, l_1) \ldots (x_n, l_n)$ where $l_i \in \{-1, +1\}$ refers to whether the patient $i$ is diagnosed with the target disease or not, the early disease detection task is to determine if a new patient's data vector $x$ would develop into the target disease by classifying the vector $x$ to $+1$ (positive) or $-1$ (negative).

## 2.2 Linear Discriminant Analysis

To solve the binary classification problem aforementioned, we consider a simple LDA classifier $f(x) \in \{\pm 1\}$ based on the given $p$-dimensional data vector $x$ and labeled samples $x_1, x_2, ...x_n$

$$f(x, \hat{\Sigma}) = sign\left((x - \bar{x})^T \hat{\Sigma}^{-1} (\bar{x}_{+1} - \bar{x}_{-1})\right) \quad (1)$$

where $\bar{x}$ refers to the mean vectors of all samples $x_1, x_2, ...x_n$; $\bar{x}_{+1}$, $\bar{x}_{-1}$ refer to the mean vectors of the positive samples and negative samples receptively.

The $\hat{\Sigma}$ is the covariance matrix estimated from data $x_1, x_2, ...x_n$. The most common estimation of $\hat{\Sigma}$ is the sample estimation:

$$\bar{\Sigma} = \frac{1}{n-1} \sum_{1 \leq j \leq n} (x_j - \bar{x})^T (x_j - \bar{x}) \quad (2)$$

Thus, we write $f(x, \bar{\Sigma})$ as the classical Fisher's Linear Discriminant Analysis.

## 2.3 Bayesian Voting Scheme

Given a binary classifier $h_\omega(x) \in \{\pm 1\}$, which is parameterized by $\omega$, the Bayesian Voting Classification [24] of the classifier is:

$$sign\left(\int_\omega h_\omega(x) p(\omega) d_\omega\right), \quad (3)$$

where the signal function $sign(\cdot)$ maps the non-negative input to $+1$ and the negative input to $-1$, and $p(\omega)$ is the prior probability of the parameter $\omega$. As a binary classifier, the above classifier in Eq. 3 outputs the label with the highest weighted vote. The theoretical advantages of Bayesian voting scheme are addressed in [24].

## 2.4 Problem Formulation

To handle the uncertainty of (inverse-) covariance matrix estimates for LDA, through combining Bayesian Voting and LDA, we can consider a new classifier as:

$$sign\left(\int_{\hat{\Sigma} \geq 0} f(\mathbf{x}, \hat{\Sigma}) P(\hat{\Sigma} | x_1, x_2, ...x_n, \mathbf{x}) d\hat{\Sigma}\right), \quad (4)$$

where $P(\hat{\Sigma}|x_1, x_2, ...x_n, \mathbf{x})$ is the probability of the covariance matrix $\hat{\Sigma}$, given the $n$ training samples $x_1, x_2, ...x_n$ as well as the new sample for prediction $\mathbf{x}$. In our research, we named this pattern as *Input Adaptive Bayesian Voting*. Note that we take the new input vector $x$ into account for generating the "hypothesis" $\hat{\Sigma}$ of Bayesian inference.

With all above backgrounds and settings in mind, the problem of this research is to compute Equation 4. However, there exists at least two major technical challenges:

***Challenge I: Fast Computation and Lazy Sampling -*** To compute the integral in Eq. 4, a common solution is to leverage a Monte-Carlo Integration algorithm [28] that first randomly samples a group of positive-semidefinite matrices e.g., $\Sigma_1, \Sigma_2 \ldots \Sigma_m$ from the distribution with probability density function $P(\hat{\Sigma}|x_1, x_2, ...x_n, \mathbf{x})$, then averages $f(\mathbf{x}, \hat{\Sigma})$ over the sampled positive-semidefinite matrices as $1/m \sum_{i=1}^{m} f(x, \Sigma_i)$. This method can give an approximate result of Eq. 4. However, the density function of the sampled positive-semidefinite matrices $P(\hat{\Sigma}|x_1, x_2, ...x_n, \mathbf{x})$ depends on the input $\mathbf{x}$. That means, for each new testing sample $\mathbf{x}$, we have to build a new probability distribution based on $P(\hat{\Sigma}|x_1, x_2, ...x_m, \mathbf{x})$, then sample a new group of positive-semidefinite matrices and run the Monte-Carlo Integration accordingly. Obviously, the computational cost to re-sample a new group of positive-semidefinite matrices for each new input $\mathbf{x}$ is high. Thus, we need a *"Lazy Sampling"* mechanism, which only samples a group of positive-semidefinite matrices once, then uses the same group of matrices for arbitrary input $\mathbf{x}$.

*Challenge II: Approximation and Convergence -* The accuracy of classification highly depends on whether the proposed algorithm can approximate to the Eq. 4 as well as the convergence rate. For the high-dimensional numeric integration [29], the approximation is usually bottle-necked by the number of dimensions (e.g., the dimensionality of positive-semidefinite matrices $p \times p$) and the sampling complexity (e.g., the number of sampled positive-semidefinite matrices $m$). Intuitively, the convergence of algorithms can be improved, with increasing sampling complexity and lower dimensionality. However, we aim at proposing an algorithm to approximate Eq. 4 with a low computational/sampling complexity while ensuring a fast convergence rate. Especially we require a convergence rate that is not sensitive to the dimensionality of the data $p$, so as to enable the high dimensional data classification.

In the rest of this paper, we present a novel classifier, *Fast Wishart Discriminant Analysis* – `FWDA`, which tackle the two research challenges, with low computational/sampling complexity and proven dimensionality-insensitive convergence rate.

## 3 `FWDA`: Algorithms and Analysis

In this section, we introduce our solution to compute Eq. 4 as follows: we first reformulate Eq. 4. Then, we introduce the algorithms of `FWDA` to compute the reformulation of Eq. 4. Finally, we analyze `FWDA`.

### 3.1 Problem Reformulation

We first define $P(x|\Sigma)$ as the probability of input vector $x$ given the covariance matrix $\Sigma$, and $P(\Sigma|x_1, x_2...x_n)$ as the probability of the covariance matrix $\Sigma$, given the training samples $x_1, x_2...x_n$. Then, we define a function:

$$g(x) = \int_{\Theta \geq 0} f(x, \Sigma) P(x|\Sigma) P(\Sigma|x_1, x_2...x_n) d_{\Sigma}. \tag{5}$$

**Theorem 1.** *Eq. 4 is equivalent to the classification result of $sign(g(x))$.*

*Proof.* Assuming all $x_1, x_2, ...x_n, \mathbf{x}$ are *i.i.d* drawn from an unknown distribution, according to the Bayesian theorem, we decompose $P(\Sigma|x_1, ...x_n, x)$ as

$$P(\Sigma|x_1, ...x_n, x) = \frac{P(x|\Sigma)P(x_1...x_n|\Sigma)P(\Sigma)}{P(x)P(x_1...x_n)} \quad (6)$$
$$= P(x|\Sigma)P(\Sigma|x_1, x_2...x_n) \cdot P(x)^{-1}$$

Thus, Eq. 4 can be re-written as $sign(p(x)^{-1} \ g(x))$. As $p(x)^{-1}$ is positive for $\forall x$. Thus, we can conclude $sign(g(x)) = sign(p(x)^{-1} \ g(x))$ should be consistently equivalent to the Eq. 4. □

Thus, the key of proposed research is to compute Eq. 5. We propose a straightforward method (FWDA): the algorithm consists of a probabilistic model that can generate $m$ sampled (inverse) covariance matrices according to the density function $P(\Sigma|x_1, x_2...x_n)$, then calculates Eq. 4 through Monte-Carlo Integration using the sampled (inverse) covariance matrices. The design of FWDA is described in the following.

### 3.2 Wishart Distribution Model based on De-sparsified Graphical Lasso

To sample (inverse) covariance matrices according to $P(\Sigma|x_1, x_2...x_n)$, FWDA leverages a Wishart Distribution [30] namely $\mathcal{W}(\hat{T}, v)$, where $\hat{T}$ refers to the "mean" positive-definite matrix for the Wishart distribution and $v$ is the degree of freedom.

Given any $p \times p$ positive definite matrix $\Theta$ (as the inverse of potential covariance matrix), we estimate the probability density of $\Theta$, based on $\mathcal{W}(\hat{T}, v)$, as:

$$P_w(\Theta|\hat{T}, v) = \frac{1}{2^{vp/2} \left|\hat{\mathbf{T}}\right|^{v/2} \Gamma_p\left(\frac{v}{2}\right)} |\Theta|^{(v-p-1)/2} e^{-(1/2)\,\mathrm{tr}(\hat{\mathbf{T}}^{-1}\Theta)} \quad (7)$$

where $|\cdot|$ refers to the determinant and the multivariate gamma function is defined as:

$$\Gamma_p\left(\frac{v}{2}\right) = \pi^{p(p-1)/4} \prod_{j=1}^{p} \Gamma\left(\frac{v}{2} + \frac{1-j}{2}\right).$$

Specifically, in our research, we set the degree of feedom $v$ as $v = n - 1$, and further estimate $\hat{T}$ using De-sparsified Graphical Lasso [31]:

$$\hat{T} = 2\hat{\Theta} - \hat{\Theta}\bar{\Sigma}\hat{\Theta}. \quad (8)$$

where $\hat{\Theta}$ refers to the Graphical Lasso estimator

$$\hat{\Theta} = \underset{\Theta \geq 0}{\mathrm{argmin}} \left( \mathrm{tr}(\bar{\Sigma}\Theta) - \log|\Theta| + \lambda \sum_{j \neq k} |\Theta_{jk}| \right), \quad (9)$$

where $\bar{\Sigma}$ refers to the sample covariance matrix on the samples $x_1, x_2, ...x_n$, $\sum_{j \neq k} |\Theta_{jk}|$ refers to the sum of absolute value of the non-diagonal elements in matrix $\Theta$.

## 3.3 Binary Classification as Bayesian Inference via Regularized Wishart Prior

Using the typical inverse-wishart sampling algorithm [32], `FWDA` first randomly generated $m$ inverse-covariance matrices $\Theta_1, \Theta_2...\Theta_m$ drawn from the Wishart Distribution $\mathcal{W}(\hat{T}, v)$. With the $\Theta_1, \Theta_2...\Theta_m$, we approximate Eq. 4 as:

$$\bar{g}(x) = \frac{1}{m} \sum_{1 \leq i \leq m} \left( f(x, \Theta_i^{-1}) P(x|\Theta_i^{-1}) \right), \quad (10)$$

where $P(x|\Theta_i^{-1})$ refers to the probability of the input vector **x** given the inverse covariance matrix $\Theta_i$. In this paper, we characterize the probability as:

$$P(x|\Theta_i^{-1}) = \frac{1}{\sqrt{2\pi|\Theta_i^{-1}|}} e^{-\frac{1}{2}(\mathbf{x}-\bar{\mathbf{x}})^T \Theta_i (\mathbf{x}-\bar{\mathbf{x}})}, \quad (11)$$

where $\bar{x} = n^{-1} \sum_1^n x_j$ refers to the mean vector of all training data.

Thus, our algorithm `FWDA` uses $sign(\bar{g}(x))$ as the classification result. The performance analysis of the proposed algorithm based on Eq. 10 to approximating the formulated problem expressed in Eq. 4 will be addressed in the following section.

## 3.4 Approximation Analysis

In this section, we present how close $\bar{g}(x)$ used in `FWDA` can approximate the reformulated problem $g(x)$.

First of all, considering the fast convergence rate of De-Sparsified Graphical Lasso [31] i.e., $||\hat{T} - \Theta^*||_\infty = \mathcal{O}_p(\sqrt{log\ p\ /n})$, with a fixed number of dimensions $p$ and an increasing number of samples $n$, we are more confident to follow an assumption frequently made in many of previous Bayesian inference studies [33, 34, 35]:

**Assumption 1.** *For any positive-semidefinite matrix $\Sigma$ i.e., $\forall \Sigma \geq 0$ and $\Theta = \Sigma^{-1}$, there exists $P(\Sigma|x_1, x_2...x_n) = P_w(\Theta|\hat{T}, v)$, where $P_w(\Theta|\hat{T}, v)$ refers to the Wishart probability of $\Theta$ based on the mean positive-semidefinite matrix $\hat{T}$ and $v = n - 1$. $\hat{T}$ is an estimate of inverse covariance matrix on samples $x_1, x_2...x_n$.*

With Assumption 1., we can substitute $P(\Sigma|x_1, x_2...x_n)$ with $P_w(\Theta|\hat{T}, v)$ i.e., the conjugate prior of inverse covariance matrix based on Wishart Distribution, to enable the Bayesian inference.

**Theorem 2.** *Under Assumption 1, for any $\eta > 0$ sufficiently small, as the number of sampled inverse covariance matrices $m \to \infty$, our algorithm $\bar{g}(x)$ converges to $g(x)$ with convergence rate $\mathcal{O}_p(\sqrt{-\log(\eta/2)/2m})$ with probability at least $1 - \eta$.*

*Proof.* Sampled inverse covariance matrices $\Theta_1, \Theta_2, ..., \Theta_m$ are i.i.d and all drawn from the Wishart distribution $\mathcal{W}(\hat{T}, v)$ with probability density function $P_w(\Theta|\hat{T}, v)$. By the classical Law of Large Numbers we know that as $m \to \infty$ we have

$$\lim_{m \to \infty} \bar{g}(x) = \int_{\Theta \geq 0} f(x, \Theta^{-1}) P(x|\Theta^{-1}) P_w(\Theta|\hat{T}, v) d\Theta = g(x),$$

under Assumption 1.

Let $\delta^2$ be the variance of $f(x|\Theta^{-1})P(x|\Theta^{-1})$ under the distribution given by $P_w(\Theta|\widehat{T}, v)$, so that

$$\begin{aligned}\delta^2 &= \mathbf{Var}_w f(x|\Theta^{-1})P(x|\Theta^{-1}) \\ &= \int_{\Theta \geq 0} \left(f(x, \Theta^{-1})P(x|\Theta^{-1}) - g(x)\right)^2 P_w(\Theta|\widehat{T}, v) d\Theta \;.\end{aligned}$$

By the Central Limit Theorem we know that for any $\gamma > 0$ we have

$$\lim_{m \to \infty} P_w \left( |\bar{g}(x) - g(x)| \leq \gamma \frac{\delta}{\sqrt{m}} \right) = \frac{1}{\sqrt{2\pi}} \int_{-\gamma}^{\gamma} e^{-t^2/2} dt \;.$$

Moreover, based on Hoeffding's inequality [Hoeffding, 1963], we can conclude that for any $\eta > 0$ sufficiently small, as $m$ is large, with probability at least $1 - \eta$ we have

$$|g(x) - \bar{g}(x)| \leq \sqrt{-\frac{1}{2m} \cdot \log\left(\frac{\eta}{2}\right)} \;.$$

□

Based on **Theorem. 2**, we can conclude that the classification result of $sign(\bar{g}(x))$ should be equivalent to Eq. 4, when the number of sampled inverse covariance matrices $m$ is large. Our later experiments show that, with more than 100 sampled inverse covariance matrices $m \geq 100$, `FWDA` can deliver decent performance and consistently outperform baseline algorithms, including SVM, Kernel SVM, Random Forest and AdaBoost.

## 4 Evaluation

In this section, we first introduce the experimental design of our evaluation. Then we present the experimental results, including the performance comparison between the proposed `FWDA` algorithm, existing LDA baselines and other predictive models. Moreover, a comparison between `FWDA` and the method using a simple discretization strategy to support our theoretical analysis of `FWDA`.

### 4.1 Experimental Setups

We use the de-identified EHR data from the College Health Surveillance Network (CHSN), which contains over 1 million patients and 6 million visits from 31 student health centers across the United States [36]. Among all diseases recorded in CHSN, we choose mental health disorders, including *anxiety disorders, mood disorders, depression disorders, and other related disorders*, as the targeted disease for early detection. We represent each patient using his/her diagnosis-frequency vector based on the clustered code set (the number of dimensions $p = 295$), where four clustered codes are considered to represent the diagnoses of mental health disorders and we do not predict these four types of mental disorders separately. Specifically, if a patient has any of these four codes in his/her EHR, we say that he/she has been diagnosed with mental health disorders as ground truth.

In order to test the EARLY detection of diseases, for each patient with mental health disorders, we use his/her historical EHR data that was generated 90 days before he/she received the first mental health disorders diagnosis. Further, patients with less than two visits were excluded from the analysis.

To demonstrate the effectiveness of our method, we compare our method with baseline algorithms in terms of the following metrics: *Accuracy* and *F1-Score*. Specifically, the Accuracy metric characterizes the proportion of patients who are accurately classified in the early detection of mental disorders. The F1-Score measures both correctness and completeness of the early detection.

**Baseline Algorithms -** To validate the superiority of `FWDA` over classical LDA, we use six baseline approaches for comparison:

- *LDA* – This algorithm is based on the common implementation of generalized Fishier's discriminant analysis listed in Equation 1. Specifically, *LDA* uses the sample covariance estimation, and inverts the covariance matrix using pseudo-inverse [37] when the matrix inverse is not available.

- *Linear Support Vector Machine (SVM-Linear)* and *SVM with Gaussian Kernels (SVM-G)* - We compare our algorithm with both linear SVM (SVM-Linear) and nonlinear Kernel SVM with Gaussian Kernels (SVM-G). Both SVM classifiers are well-tuned among a wide range of parameters. Specifically, we report the performance of SVM-G classifiers with bandwidth parameter 0.1 and 1.0 in our research.

- *Decision Tree (D-Tree)*, *Random Forest* and *AdaBoost* - To compare our solution with the tree-based hypotheses, we use Decision Tree (D-Tree), and Random Forest for comparisons. Further, an AdaBoost classifier based on Logistic Regression is also used for comparison, as an advanced logistic regression baseline. Specifically, we report the performance of Random Forest and AdaBoost with 100 and 200 classifiers instances, respectively.

- *Two-stage LDA*, *Logistic Regression* and *LDA with Shrinkage Estimators* - We also compared `FWDA` to other competitors including two-stage LDA [38, 39], logistic regression [40], and LDA with shrinkage estimators [12].

We perform experiments with the following settings: to build the *training sets*, we randomly select 50 to 500 patients with mental health disorders as the positive training samples, and randomly select the same number of patients not been diagnosed with any mental health disorders as negative training samples. Thus the training set for the two classes is balanced (i.e., the number of dimensions $p = 295$ and training set size is $50 \sim 500 \times 2$). To build the testing sets, we randomly select 200 patients (not included in the training set) from both positive/negative groups. Also the testing set is balanced. For each setting, we execute the seven algorithms and repeat 30 times.

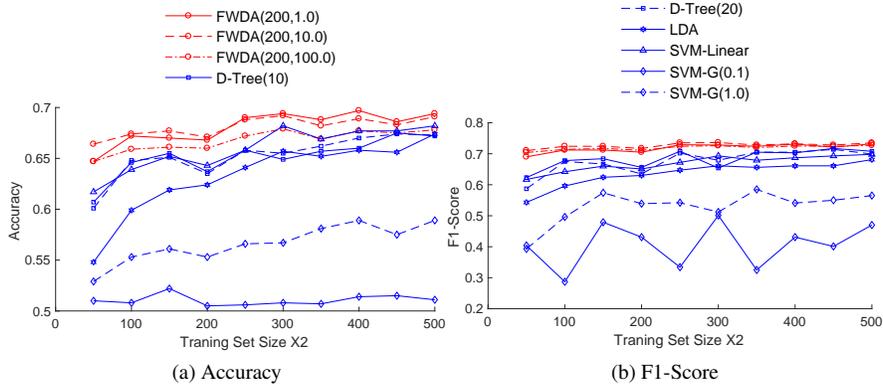

Figure 1: Overall Performance Comparison with Downstream Classifiers and Regularized LDA

## 4.2 Experimental Results

### 4.2.1 Overall Comparison

Figure. 1 presents the performance of our approach, along with classical LDA, linear SVM, Kernel SVM and Decision Trees on 200 testing samples and varying training sample sizes. FWDA(200, 1.0) refers to the FWDA classifier based on 200 sampled inverse covariance matrices using De-Sparsified Graphical Lasso with $\lambda = 1.0$ for Wishart mean matrix estimation. As can be seen from the results, FWDA clearly outperforms the baseline algorithms in terms of overall accuracy, and F1-score.

Due to the space limit, we don't present the comparison results based on Two-stage LDA, Logistic Regression and LDA with shrinkage estimators in Figure. 1. FWDA(200, 1.0) outperforms Two-stage LDA by achieving on average 4.3% higher accuracy and 3.5% higher F1-score, it outperforms Logistic Regression with, on average, 13.2% higher accuracy and 70.4% higher F1-score, and also outperforms LDA with shrinkage estimators (with fined-tuned parameters) with, on average, 12.5% higher accuracy and 23.9% higher F1-score. It is clear that FWDA outperforms all these algorithms significantly.

### 4.2.2 Comparison to Ensemble learners

As FWDA ensembles the classification results from multiple classifiers, we also compared FWDA to the existing ensemble learning algorithms, such as Random Forest and AdaBoost. To compare with ensemble learners with 100 and 200 basis classifiers, we use FWDA with 100 and 200 sampled inverse covariance matrices (i.e., ensemble with 100 and 200 LDA classifiers), with $\lambda = 1.0$ for Wishart mean matrix regularization.

The performance comparison is illustrated in Figure. 2. It is obvious that FWDA outperforms these two algorithms in both 100-instance and 200-instance settings, while the performance of Random Forest is not quite stable. Moreover, Figure. 2 also shows the performance of FWDA classifiers with 100 and 200 sampled inverse covariance matrices are very similar. Indeed, we tested FWDA with 50 to 2000 inverse covariance matrices, the prediction accuracy or F1-scores of FWDA is almost consistent on the varying number of sampled matrices. This indicates that FWDA can provide robust prediction performance, even when only a small number of inverse covariance matrices are sampled.

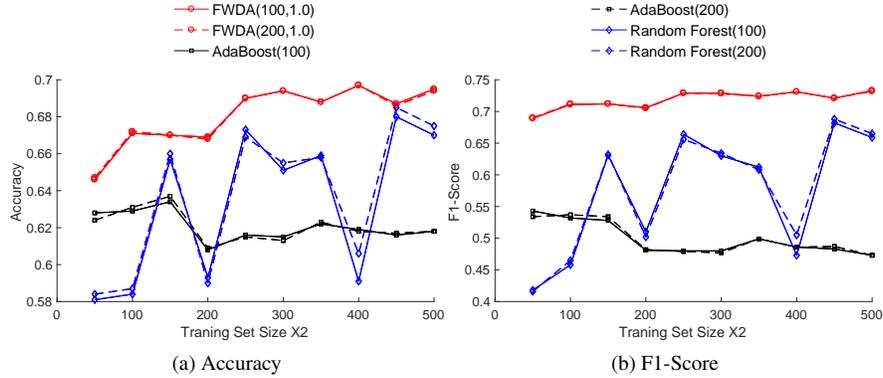

| (a) Accuracy | (b) F1-Score |

Figure 2: Performance Comparison with Ensemble Learning Classifiers

#### 4.2.3 Comparison on Discretization and Regularization

FWDA leverages importance sampling-alike method to improve the performance of approximation to the real integral in a discretization manner. We compare FWDA to a classifier namely "Discrete-FWDA" based on the simple discretization strategy:

$$sign\left(\sum_{1\leq i\leq m} f(x,\Theta_i^{-1})P(x|\Theta_i^{-1})P_w(\Theta_i|\widehat{T},v)\right).$$

Both two algorithms leverage a De-sparsified Graphical Lasso with $\lambda = 1.0, 10.0, 100.0$ for the Wishart mean matrix estimation. Figure. 3 presents the performance comparison. It shows FWDA outperforms the simple discretization strategy using the same $\lambda$ significantly.

In Figure 3, we also demonstrate the performance improvement contributed by regularization (De-sparsified graphical lasso) for Wishart mean matrix estimation. The lines entitled "Sample-FWDA" refer to a derived method using the sample inverse covariance matrix (pseudo-inverse when the covariance matrix is singular) as the Wishart mean matrix. It shows that FWDA can outperform Sample-FWDA significantly.

### 4.3 Comparison on Time Consumption

In addition to the accuracy comparison, we also compare the time consumption of FWDA (based on "Lazy Sampling") with an alternative method that is based on the same regularized Wishart distribution. For each new data classification, it samples a group of inverse covariance matrices for LDA classification and voting. Given $400\times 2$ training samples, FWDA(200,1.0), it on average consumes 149.5 seconds to train the model (including De-sparsified Graphical Lasso with $\lambda = 0.1$ and sampling 200 inverse covariance matrices), and 0.037 seconds to classify 200×2 samples for testing. We evaluate the alternative algorithm in the same settings – the alternative algorithm doesn't need to be trained, and it on average consumes 16.7 hours to classify the 200×2 testing samples. The time consumption to train a FWDA model is almost the same as the time consumed to classify one sample by the alternative method. The experiments were all carried out using an iMac desktop with 3.1 GHz Intel Core i5 CPU, 16G memory and macOS v10.12.

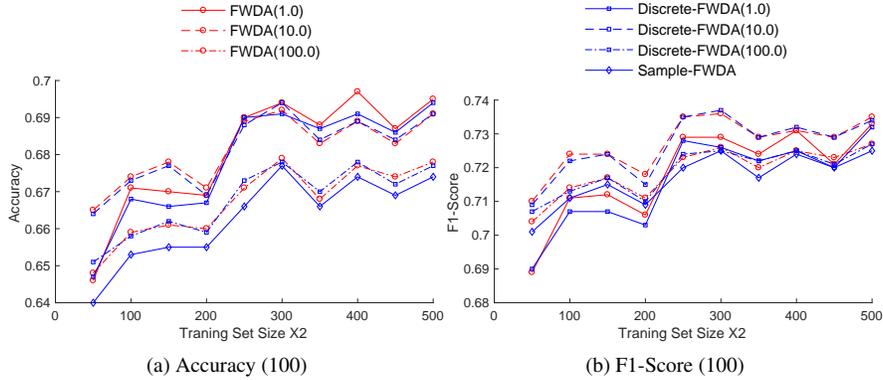

Figure 3: Performance Comparison with Different Discretization and Regularization

## 5  Conclusion and Discussion

In this paper, we proposed FWDA – a novel algorithm for early detection of diseases, based on EHR data and the diagnosis-frequency vector data representation. FWDA lowers the uncertainty of LDA parameter estimation and further enables the nonlinear classification, through the Input-Adaptive Bayesian voting scheme. The theoretical analysis shows that FWDA converges to the optimal Bayesian voting in a fast rate. The experimental results on real-world EHR dataset CHSN show that FWDA outperforms all baseline algorithms. In our future work, we intend to integrate FWDA with advanced EHR data representation techniques [1, 3, 2, 4, 6], to enable the superior prediction.